\documentclass[runningheads]{llncs}
\usepackage[T1]{fontenc}
\usepackage{caption}
\usepackage{amsmath}
\usepackage{amssymb}
\usepackage[figuresright]{rotating}
\usepackage{hyperref}
\usepackage{subfigure}
\usepackage{multirow}
\usepackage{mathrsfs}
\usepackage{makecell}
\usepackage{color}
\begin{document}
\title{Multi-modal Speech Emotion Recognition via Feature Distribution Adaptation Network}
\titlerunning{Feature distribution Adaptation Network for Speech Emotion Recognition}
%
\author{
Shaokai Li\inst{1} \and
Yixuan Ji\inst{2} \and
Peng Song\inst{1} \and
Haoqin Sun\inst{3} \and
Wenming Zheng\inst{4}
}
%
\authorrunning{S. Li et al.}
%
\institute{
School of Computer and Control Engineering, Yantai University, Yantai 264005, China\\
\and
Business School, Xuzhou University of Technology, Xuzhou 221018, China\\
\and
College of Computer Science, Nankai University, Tianjin 300350, China\\
\and
Key Laboratory of Child Development and Learning Science of Ministry of Education, Southeast University, Nanjing 210096, China
}
\maketitle              
\begin{abstract}
In this paper, we propose a novel deep inductive transfer learning framework, named feature distribution adaptation network, to tackle the challenging multi-modal speech emotion recognition problem. Our method aims to use deep transfer learning strategies to align visual and audio feature distributions to obtain consistent representation of emotion, thereby improving the performance of speech emotion recognition. In our model, the pre-trained ResNet-34 is utilized for feature extraction for facial expression images and acoustic Mel spectrograms, respectively. Then, the cross-attention mechanism is introduced to model the intrinsic similarity relationships of multi-modal features. Finally, the multi-modal feature distribution adaptation is performed efficiently with feed-forward network, which is extended using the local maximum mean discrepancy loss. Experiments are carried out on two benchmark datasets, and the results demonstrate that our model can achieve excellent performance compared with existing ones. Our code is available at https://github.com/shaokai1209/FDAN.
\end{abstract}

\section{Introduction}
\label{sec:introduction}
Speech emotion recognition (SER) aims at mining human's emotional information from speech signals, and has attracted extensive attention due to its wide practical applications \cite{singh2022systematic,vettoruzzo2024advances}. In the past dacades, many machine learning methods have been applied to SER, e.g., support vector machine (SVM), Gaussian mixture model (GMM), least square regression, subspace learning, and deep learning methods \cite{singh2022systematic,latif2023survey}. These previous works have achieved satisfactory results in the SER task, but the emotion expression ability of the single modal feature is limited, and more involved modalities would boost the recognition performance. In this way, the challenging task for SER is referred to as multi-modal SER. 
\par
To tackle the above challenging task, some methods have tried to introduce linguistic information to improve the results of SER \cite{liu2022multi,yang2022contextual,wang2023exploring,khan2023mser}. In \cite{liu2022multi}, Liu et al. use self-attention convolutional neural network (CNN) and self-attention long-short term memory (LSTM) to learn the multi-scale fusion features of text and speech, respectively, and then improves results of multi-modal SER through a decision fusion layer. In \cite{yang2022contextual}, Yang et al. develop a cross-modal transformer to explore the interactions between text and audio modalities to improve the results of SER. In \cite{wang2023exploring}, Wang et al. propose a modality-sensitive multi-modal speech emotion recognition framework to deal with this challenging task. In \cite{khan2023mser}, Khan et al. propose a deep feature fusion technique using a multi-headed cross-attention mechanism. However, according to $7\%-38\%-55\%$ criterion for emotion expression \cite{mehrabian2017communication}, the emotional expression of linguistic information is far inferior to that of facial expression information. Therefore, in this paper, we focus on facial expression information and acoustic information to complete the challenging multi-modal SER task \cite{ito2021audio,wei2022audio}. 
\par
Over the past decades, transfer learning has proven to be an effective strategy to improve the generalization ability of emotion classification \cite{poria2017review,hazarika2021conversational,vettoruzzo2024advances}.  Theoretically, different modals can be regarded as different domains, so that the multi-modal emotion recognition task can be solved by a transfer learning framework. Recently, many studies have also successfully introduced transfer learning into multi-modal emotion recognition frameworks \cite{yang2022disentangled,sharma2023real,ma2023transformer,ghorbanali2024capsule}, which can effectively improve the emotion recognition performance. 
\par
However, the existing multi-modal classification methods do not fully consider the discrepancy in feature distribution between different modalities. Thus, different from previous multi-modal SER frameworks, in this work, we introduce the deep transfer learning framework for multi-modal SER, aiming to improve the generalization ability and recognition performance of emotion model by aligning the multi-modal feature distributions. According to \cite{pan2009survey}, due to both modal feature domains of our framework are guided by label information  during the training process, so our work belongs to the category of inductive transfer learning.
\par
\begin{figure}[t]
\centering
\includegraphics[scale=0.3]{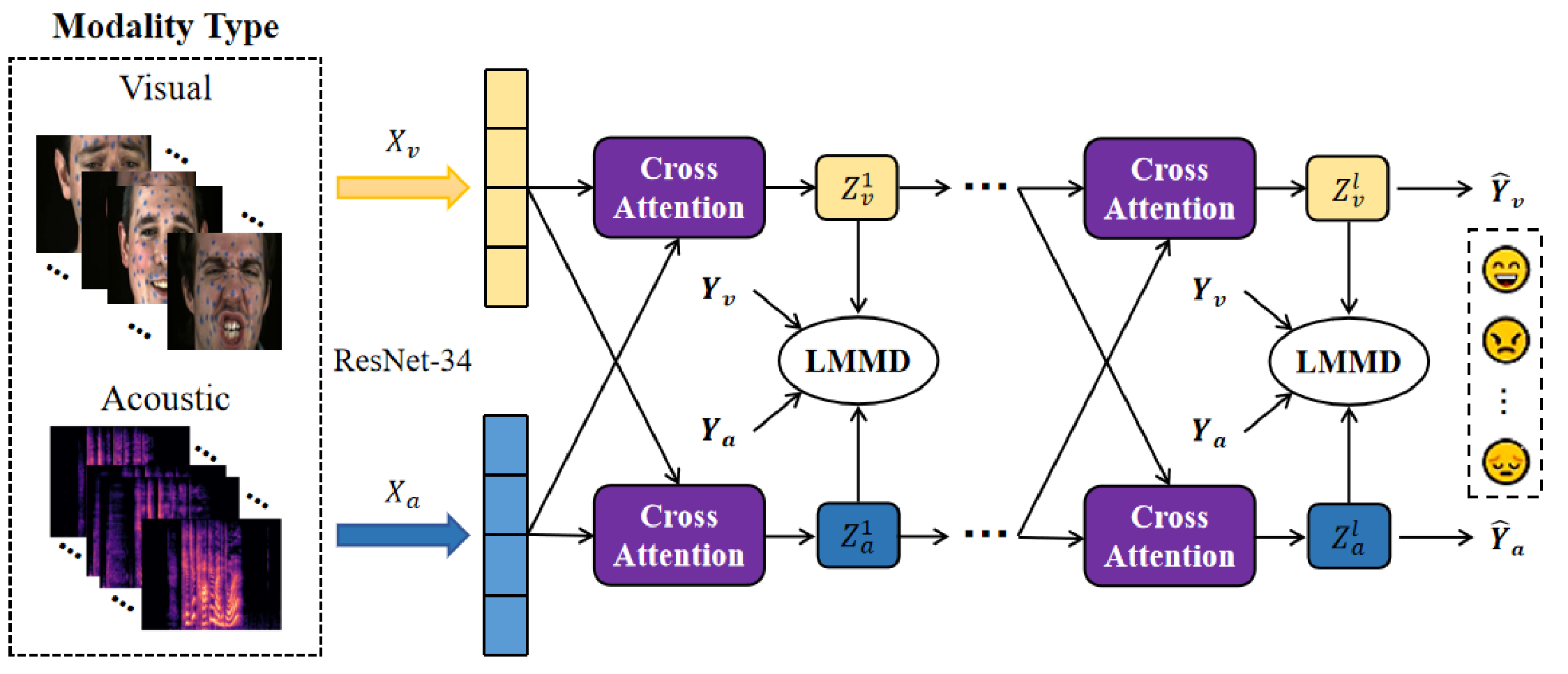}
\captionsetup{font={small}}
\caption{The framework of FDAN. The blue and yellow parts represent the learning process of visual and acoustic features respectively. The purple part represents the cross-attention mechanism. $X_v$ and $X_a$ represent visual and acoustic features extracted by the pre-trained ResNet-34 respectively. By minimizing the LMMD loss, the feature distribution discrepancy between the coupled feature subspaces $Z_v^i$ and $Z_a^i$ in the $i$-th layer is reduced, where $i\in l$. $Y_v$ and $Y_a$ represent  the true labels for the visual and acoustic samples, and $\hat{Y}_v$ and $\hat{Y}_a$ represent the corresponding predicted labels.}
\label{fig:flowchart}
\end{figure}

In this paper, we propose a novel deep transfer learning framework called feature distribution adaptation network (FDAN) for multi-modal SER tasks (see Fig. \ref{fig:flowchart}). Specifically, first, the pre-trained ResNet-34 \cite{he2016deep} on ImageNet is selected as the feature extraction for visual and speech modalities, respectively. Second, the cross-attention mechanism is introduced to model the intrinsic similarity relationships of multi-modal features. Third, the multi-modal feature distribution adaptation is achieved efficiently with feed-forward network by extending it with local maximum mean discrepancy (LMMD) loss \cite{zhu2020deep}. It is worth mentioning that the proposed model focuses on narrowing the feature distribution among different modalities. Compared to the existing multi-modal SER methods, the proposed method does not need to perform association learning on different modal features of the same sample. To verify the effectiveness of the proposed framework, we conduct extensive experiments on two popular datasets, and the experimental results indicate that the proposed framework can effectively improve the results of multi-modal SER tasks. 

\section{The proposed method}
\label{sec:proposed method}
\subsection{Preliminary}
Following the symbolic definition of transfer learning, given a visual domain $\mathcal{D}_v = {(x_{vn},y_{vn})}_{n=1}^{n_v}$ and a acoustic domain $\mathcal{D}_a = {(x_{an},y_{an})}_{n=1}^{n_a}$, where $x_{vn}$ and $x_{an}$ represent samples in the visual and acoustic domains respectively, $y_{vn}$ and $y_{an}$ correspond to their labels respectively, and $n_v$ and $n_a$ represent the numbers of samples. $p_v$ and $p_a$ represent feature distributions in the visual and acoustic domains. $p_a^{(c)}$ and $p_v^{(c)}$ represent the feature distribution of samples belonging to class $c$ in the visual and acoustic domains. $X_v$ and $X_a$ represent the visual and acoustic feature matrices. We define the feature subspace of the visual modality in the $i-$th layer as $Z_v^i\in \textbf{R}^{n_v\times d_v}$ and the feature subspace of the acoustic modality in the $i-$th layer as $Z_a^i \in \textbf{R}^{n\times d_a}$. The loss of our model is formulated as
\begin{equation}
\begin{aligned}
& \min_f \frac{1}{n_a}\sum_{n=1}^{n_a}\mathcal{J}(f(x_{an}),y_{an}) + \frac{1}{n_v}\sum_{n=1}^{n_v}\mathcal{J}(f(x_{vn}),y_{vn}) + \alpha \textbf{E}_c[d(p_v,p_a)] 
\end{aligned}
\label{eq:model_loss}
\end{equation}
where $\mathcal{J}(\cdot ,\cdot)$ is the cross-entropy loss function, $d(\cdot ,\cdot)$ is the domain adaptation loss, $\textbf{E}_c[\cdot]$ is the mathematical expectation of the class, $f(\cdot)$ is the consistent representation of emotion, and $\alpha$ is a trade-off parameter to balance the cross-entropy loss and the domain adaptation loss.
\subsection{Cross-attention module}\label{sec:cross-attention}
To obtain the feature correlation between the visual and acoustic modalities, the features of each modality are first projected into three feature subspaces, known as query, key, and value. The process can be computed as follows:
\begin{equation}
 Q_v = W_v^Q {Z_v^i}^T, \quad
 K_v = W_v^K {Z_v^i}^T, \quad
 V_v = W_v^V {Z_v^i}^T
\label{eq:QKV}
\end{equation}
where $Q_v$, $K_v$, $V_v \in \textbf{R}^{d_v \times n_v}$ are the query, key, and value of the visual modality, and $W_v^K$, $W_v^Q$, $W_v^V\in \textbf{R}^{d_v \times d_v}$ are the projection matrices of them. $n_v$ and $d_v$ represent the number and the feature dimension of samples of the visual modality respectively. The above calculation process can also yield the query, key and value, i.e., $Q_a$, $K_a$, $V_a \in \textbf{R}^{d_a \times n_a}$ of the acoustic modality, and their associated projection matrices $W_a^K$, $W_a^Q$, $W_a^V\in \textbf{R}^{d_a \times d_a}$. Note that $d_a = d_v$.

\begin{figure}[t]
\centering
\includegraphics[scale=0.26]{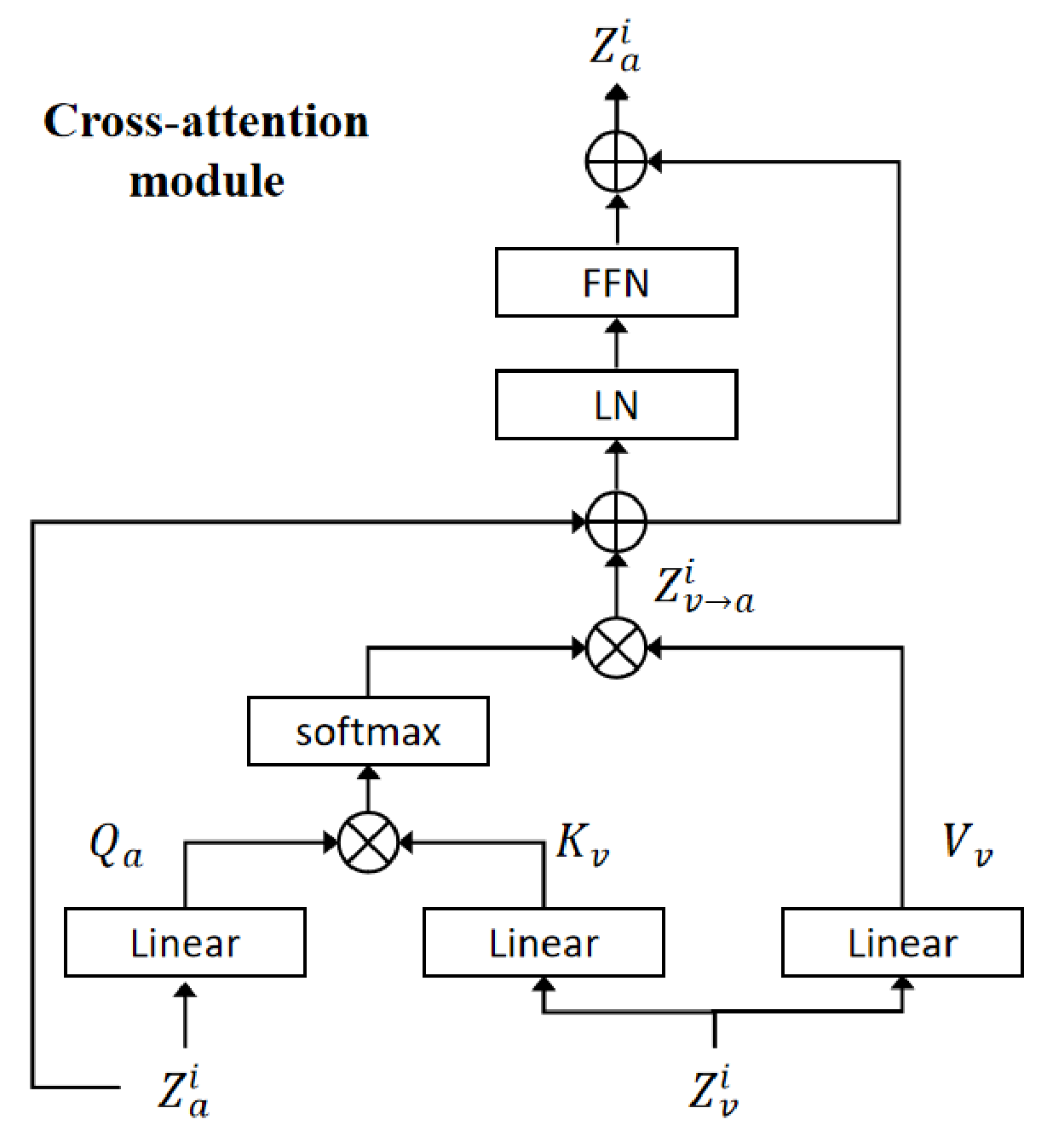}
\captionsetup{font={small}}
\caption{The structure of the cross-attention module.}
\label{fig:cross_attention}
\end{figure}
Following \cite{sun2021multimodal}, we first cross calculate the dot product of the query and key for visual and acoustic to obtain the association between the coupled modalities. The results are then scaled and normalized by the softmax function to obtain attention weights. Then, we use the corresponding weights to aggregate the value items of each feature sequence. The calculation formulas are as follows:
\begin{equation}
\begin{aligned}
& \Delta Z_{v\rightarrow a}^i =softmax(Q_a^TK_v/ \sqrt{d_v})V_v^T\\
& \Delta Z_{a\rightarrow v}^i =softmax(Q_v^TK_a/ \sqrt{d_a})V_a^T
\end{aligned}
\label{eq:cross-attention_1}
\end{equation}
where $\Delta Z_{v\rightarrow a}^i \in \textbf{R}^{n_a \times d_v}$ and $\Delta Z_{a\rightarrow v}^i \in \textbf{R}^{n_v \times d_a}$ are the propagated information of visual-to-audio and audio-to-visual, respectively. 
\par
We further update the features of one modality based on the propagation information of the other modality, the update rules are formulated as follows:
\begin{equation}
\begin{aligned}
& Z_v^i = LN(Z_v^i + \Delta Z_{a\rightarrow v}^i), \quad Z_v^i = LN(Z_v^i + FFN(Z_v^i))\\
& Z_a^i = LN(Z_a^i + \Delta Z_{v\rightarrow a}^i), \quad Z_a^i = LN(Z_a^i + FFN(Z_a^i))
\end{aligned}
\label{eq:cross-attention_2}
\end{equation}
where $LN(\cdot)$ represents the layer normalization, and $FFN$ represents the feed-forward network.
\par
Finally, the intrinsic similarity relationships of the different modalities propagate to each other in the feature subspace $Z_j^i$, $j\in \{a,v\}$. In the case of $Z_a^i$, the details of the cross-attention module are shown in Fig. \ref{fig:cross_attention}.
\subsection{Feature distribution adaptation}
\label{sec:feature-distribution-adaptation}
In order to obtain more plentiful consistent representation of emotion, we hope to align the feature distributions of the different modalities within the same category. As a popular distance metric, maximum mean discrepancy (MMD) is widely used to measure the discrepancy of the probability distribution between feature domains, so as to improve the generalization ability of emotion classification models \cite{li2022transferable} \cite{lu2022domain}. Unlike traditional MMD, this paper follows \cite{zhu2020deep}, but uses true labels to guide MMD learning. While aligning the feature distribution of different modalities, the feature distribution of samples belonging to the same label is more compact. This process is called local maximum mean discrepancy (LMMD). The objective function of LMMD is written as follows:
\begin{equation}
\begin{aligned}
& d_{\mathcal{H}}(p_v,p_a) \triangleq \textbf{E}_c\|\textbf{E}_{p_v^{(c)}}[\phi (X_v)] - \textbf{E}_{p_a^{(c)}}[\phi (X_a)]\|_{\mathcal{H}}^2
\end{aligned}
\label{eq:lmmd}
\end{equation}
where $p_a^{(c)}$ and $p_v^{(c)}$ represent the feature distribution of samples belonging to class $c$ in the visual and acoustic domains. $\mathcal{H}$ is the  reproducing kernel Hilbert space (RKHS), and $\phi (\cdot)$ represents its mapping function, which is usually implemented based on a kernel function. We use the weight $w_c$ to determine which category each sample belongs to, and Eq. (\ref{eq:lmmd}) can be further transformed into the following form:
\begin{equation}
\begin{aligned}
& d_{\mathcal{H}}(p_v,p_a) \triangleq \frac{1}{C}\sum_{c=1}^C \| \sum_{x_{vn}\in X_v}w_{vnc}\phi (x_{vn}) -\!\!\!\!\! \sum_{x_{am}\in X_a} \!\!\!\! w_{amc} \phi (x_{am})\|_{\mathcal{H}}^2
\end{aligned}
\label{eq:w_lmmd}
\end{equation}
where $w_{vnc}$ and $w_{amc}$ represent the weight of $x_{vn}$ and $x_{am}$ belonging to class $c$, and $m\vee n \in n_a\vee n_v$. $w_{jnc}$ ($j\in \{a,v\}$) can be computed as follows:
\begin{equation}
\begin{aligned}
& w_{jnc} = \frac{y_{jnc}}{\sum_{(x_{jm},y_{jm})\in \mathcal{D}_{j}}y_{jmc}}
\end{aligned}
\label{eq:w_weight}
\end{equation}
where $y_{jnc}$ is the $c-$th entry of vector $y_{jn}$. Obviously, the labels $y_{vn}$ and $y_{am}$ are the one-hot vectors, which are used to compute $w_{vnc}$ and $w_{amc}$ in equation (\ref{eq:w_lmmd}).
\par
Note that we cannot directly calculate the $\phi(\cdot)$ in Eq. (\ref{eq:w_lmmd}). To align the feature distributions of the coupled modalities, the network will generate  the activation after the cross-attention module in the $i-$th layers as $\{{z_v^i}_n\}_{n=1}^{n_v}$ and $\{{z_a^i}_n\}_{n=1}^{n_a}$, $i\in l=\{1,2,...,l\}$. Thus,  Eq. (\ref{eq:w_lmmd}) can be reformulated as follows:
\begin{equation}
\begin{aligned}
& d_i(p_v, p_a)= \frac{1}{C}\sum_{c=1}^C \bigg[ \sum_{m=1}^{n_v}\sum_{n=1}^{n_v}w_{vmc} w_{vnc} k(z_{vm}^i,z_{vn}^i) \\
&\!+ \! \sum_{m=1}^{n_a}\! \sum_{n=1}^{n_a}w_{amc} w_{anc} k(z_{am}^i,z_{an}^i)\! -\! 2 \! \sum_{m=1}^{n_v}\! \sum_{n=1}^{n_a}w_{vmc} w_{anc} k(z_{vm}^i,z_{an}^i) \! \bigg]
\end{aligned}
\label{eq:distri_adap}
\end{equation}
where $ k(z_v,z_a) = \langle\phi (z_v), \phi (z_a) \rangle $ is the kernel function, in which $\langle \cdot, \cdot \rangle $ represents the inner product of two vectors.
\par
Finally, the feature distribution adaptation loss of our model is written as follows:
\begin{equation}
\begin{aligned}
& \min_f \frac{1}{n_a}\sum_{n=1}^{n_a}\mathcal{J}(f(x_{an}),y_{an}) + \frac{1}{n_v}\sum_{n=1}^{n_v}\mathcal{J}(f(x_{vn}),y_{vn}) + \alpha \sum_{i\in l}d_i(p_v, p_a)
\end{aligned}
\label{eq:final_loss}
\end{equation}
where $f(\cdot)$ is the consistent representation of emotion, $\alpha$ is a trade-off parameter.
\section{Experiments}
\subsection{Datasets}
The following two datasets are used in our experiments. 
\begin{itemize}
\item SAVEE \cite{haq2009speaker}: It is an acted English audio-visual emotional dataset, which consists of 480 utterances from four male actors in seven emotional categories.
\item  MELD \cite{poria2018meld}: It is an acted English audio-visual-text emotional dataset, which consists of 13,708 utterances from 1,433 dialogues from TV-series $Friends$ in seven emotional categories.
\end{itemize}
Note that the samples of all emotional categories (anger, disgust, fear, happiness/joy, neutral, sadness, surprise) from the above-mentioned two datasets are used in our experiments.
\begin{figure}[t]
\centering
\includegraphics[scale=0.3]{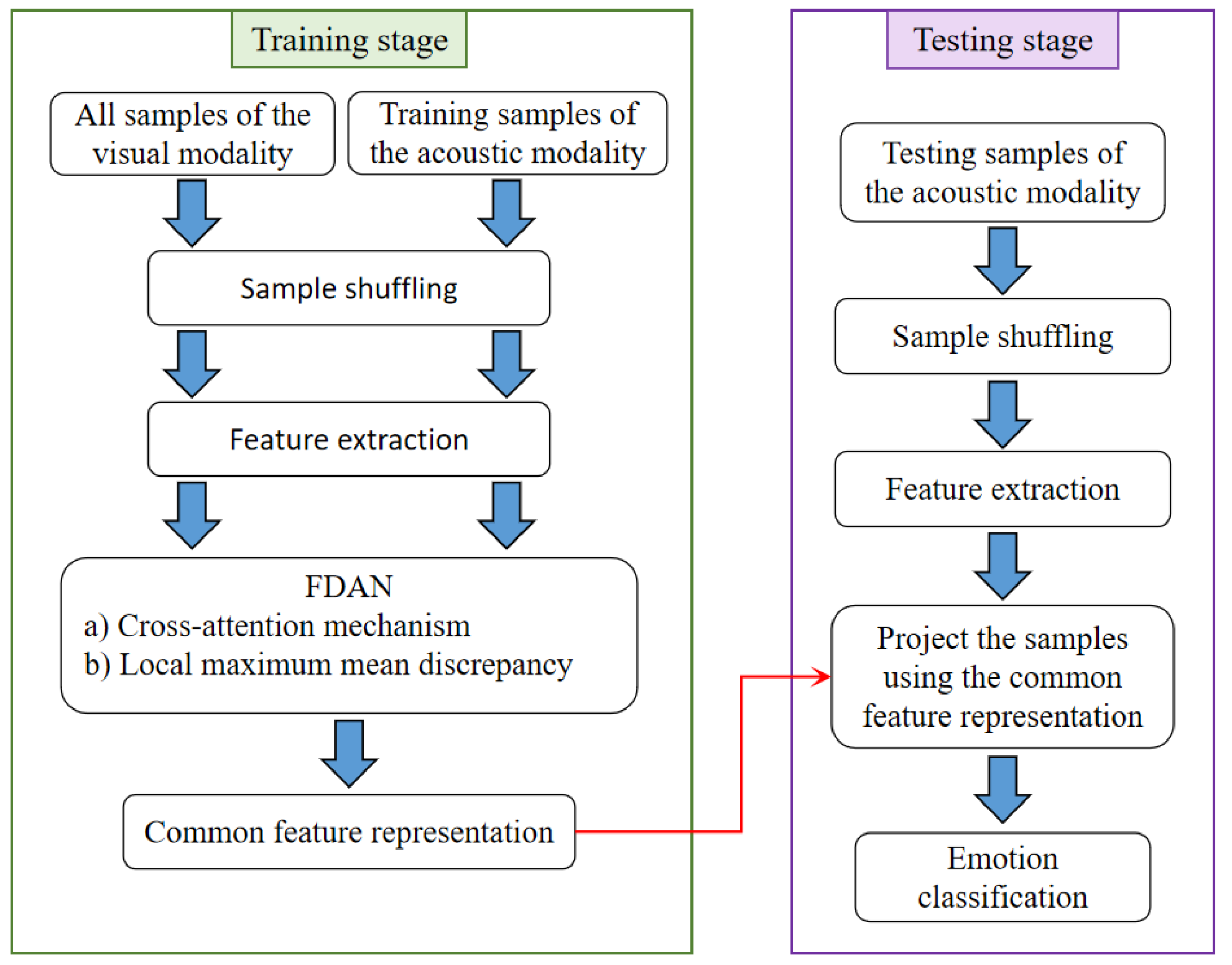}
\captionsetup{font={small}}
\caption{The framework of the training and testing process of the FDAN model.}
\label{fig:train_framework}
\end{figure}
\subsection{Experimental setup}
In our experiments, following the experimental settings of our previous works \cite{li2023generalized,wang2023common}, the visual modality is taken as the source domain, the acoustic modality is taken as the target domain, all samples of the source domain and a part of the target domain are taken for training, and the remaining samples of the target domain are taken for testing. More details are shown in Fig. \ref{fig:train_framework}. We extract the Mel spectrograms as the features of the acoustic signals, and the video frames are uniformly sampled in $\{ V_1, V_2, V_3\}$ to obtain more facial expression samples.
\par
For the division of training and test sets, we selected 8/10 speech samples of each emotional categories and all visual samples for training, and the remaining 2/10 speech samples for testing. In all experiments, the value of momentum is set to 0.99. We fine-tune the value of $\alpha$ in $\{ 10^{-3}, 10^{-4}, 10^{-5}\}$ and fine-tune the value of decay in $\{ 10^{-3}, 10^{-4}\}$. The batch size is 32, and 300 epochs are executed in the training.  We use the weighted average recall (WAR), unweighted average recall (UAR), and weighted F1-score (w-F1) as the experimental evaluation metrics. Note that the computer used in our experiments has a GeForce RTX 2080 Ti GPU and a 40 GB RAM, and the software environment is PyTorch 1.10.0.
\subsection{Baselines}
To assess the performance of our model, we compare it with the following pre-trained conventional networks, state-of-the-art multi-feature fusion and multi-modal SER methods. 
\begin{itemize}
    \item Pre-trained models \cite{shwartz2022pre}: We fine-tune the pre-trained VGG-16, VGG-19, ResNet-18, ResNet-34, and ResNet-50 in the Torchvision library.
    \item INCA \cite{tuncer2021automated}: A feature selection method for SER, which uses iterative neighborhood component analysis to select discriminative features.
    \item GM-TCNet \cite{ye2022gm}: A multi-scale feature fusion method for SER, which uses multi-scale receptive field to obtain emotional causality representation.
    \item SMIN \cite{lian2022smin}: A multi-modal emotion recognition method, which combines a semi-supervised learning framework with multi-modal emotion recognition.
    \item TRIN \cite{dong2022temporal}: A multi-modal SER method, which explores the underlying associations between different modal features under the sequential temporal guidance.
    \item SDT \cite{ma2023transformer}: A multi-modal emotion recognition method, which introduces transformer-based model with self-distillation to transfer knowledge.
    \item CADF \cite{khan2023mser}: A multi-modal SER method, which uses the multi-headed cross-attention mechanism to fuse multi-modal features.
    \item TF-Mix \cite{wang2024design}: A feature fusion method for SER, which amalgamates various feature extraction techniques to enhance emotion recognition.
    \item MAP \cite{liang2024multi}: A multi-modal emotion recognition method, which presents a multi-modal attentive prompt learning framework to improve emotion recognition in conversations.
\end{itemize}
Since the experimental settings of above methods are similar to ours and most of them do not have publicly available source code, we present and compare the reported results from their publications.
\subsection{Experimental results}
\renewcommand\arraystretch{1.1} 
\begin{table}[t]
\centering
\captionsetup{font={small}}
\caption{The recognition results of different models on the SAVEE dataset.}
\begin{tabular}{cp{2.4cm}<{\centering} cp{2.4cm}<{\centering} cp{2.4cm}<{\centering} cp{2.4cm}<{\centering}}
\hline
Models & WAR(\%) & UAR(\%) & w-F1(\%) \\ \hline \hline
VGG-16 & 64.58 & 63.33 &  64.30\\
VGG-19 & 67.70 & 66.41 & 67.71\\ 
ResNet-18 & 59.37 & 59.01 & 58.74\\ 
ResNet-34 & 69.79 & 70.33 & 70.01\\
ResNet-50 & 68.75 & 68.31 & 68.42\\
INCA \cite{tuncer2021automated}& 81.63 & 79.73 & 80.59 \\ 
GM-TCNet \cite{ye2022gm} & 84.79 & 83.33 & ---\\ 
TF-Mix \cite{wang2024design} & 86.23 & 84.71 & 84.85\\ 
Ours & \textbf{86.66} &  \textbf{86.19} & \textbf{86.96} \\\hline
\end{tabular}
\label{tab:SAVEE_results}
\end{table}

\begin{table}[t]
\centering
\captionsetup{font={small}}
\caption{The recognition results of different models on the MELD dataset.}
\begin{tabular}{cp{2.4cm}<{\centering} cp{2.4cm}<{\centering} cp{2.4cm}<{\centering} cp{2.4cm}<{\centering}}
\hline
Models & WAR(\%) & UAR(\%) & w-F1(\%) \\ \hline \hline
VGG-16 & 50.31 & 49.51 &  49.03\\
VGG-19 & 54.53 & 54.08 & 53.33\\ 
ResNet-18 & 58.68 & 59.80 & 58.41\\ 
ResNet-34 & 63.82 & 63.10 & 62.91\\
ResNet-50 & 61.02 & 60.31 & 60.79\\
SMIN \cite{lian2022smin} & 65.59 & ---& 64.50 \\ 
TRIN \cite{dong2022temporal} & 79.60 & 70.52 & ---\\ 
SDT \cite{ma2023transformer} & 67.55 & --- & 66.60\\ 
CADF \cite{khan2023mser} & --- & 72.30 & ---\\
MAP \cite{liang2024multi} & 78.50 & \textbf{78.50} & 78.40\\
Ours & \textbf{81.16} & 75.91 & \textbf{80.72} \\\hline
\end{tabular}
\label{tab:MELD_results}
\end{table}
The experimental results are shown in Tables \ref{tab:SAVEE_results} and \ref{tab:MELD_results}. From the tables, we have the following  observations:
\par
First, the performance of multi-modal and feature fusion methods in SER is significantly better than that of traditional residual networks. In addition, among VGG and ResNet networks, ResNet-34 achieves the best recognition performance, which also indicates that ResNet-34 is more suitable as the feature extractor of our model.
\par
Second, from Table \ref{tab:SAVEE_results}, we can find that our model achieves better results than INCA, GM-TCNet, and TF-Mix on the SAVEE dataset. The reason is that INCA only considers the feature selection, while GM-TCNet and TF-Mix consider feature fusion but only focuses on the acoustic modality. Moreover, from Table \ref{tab:MELD_results}, we can find that our model achieves better performance compared to SMIN, TRIN, SDT, CADF, and MAP on the MELD dataset. The reason is that SMIN considers semi-supervised multi-modal feature learning and ignores some label information. Meanwhile, although TRIN, SDT, and CADF fully consider multi-modal label information, they only focus on the relationship between different modal features of the same sample, ignoring the feature distribution between different modalities. In addition, the MAP method obtain good recognition results by introducing the prompt learning strategy to fine-tune the pre-trained language model, learning the weighted feature fusion across text, visual, and speech modalities, which outperforms the proposed method on the w-F1 metric. However, MAP did not consider the discrepancy in feature distributions between modalities, resulting in poor overall performance compared with the proposed method.
\par
Third, compared with the baseline models, our method achieves better recognition results on both the SAVEE and MELD datasets. This demonstrates the validity of our model. The reason is that, in our model, the transfer learning strategy is introduced to align the multi-modal feature distribution, which can obtain a consistent emotion representation to improve the results of SER.
\par
Last, it is worth mentioning that compared to the existing multi-modal SER methods, the proposed method does not need to perform association learning on different modal features of the same sample. Thus, our method can use cross-dataset or multi-dataset facial expression features to guide the learning of speech emotional feature representation. The results of the multi-dataset multi-modal SER are shown in Table \ref{tab:cross-corpus}. Specifically, the task SAVEE$\{S_v, M_v, S_a\}$ uses all visual samples from the SAVEE and MELD datasets and 8/10 speech samples from the SAVEE for training, and the remaining 2/10 speech samples for testing. The results show that FDAN  performs well for the multi-dataset multi-modal SER tasks. Compared with the experimental settings of the single dataset (see Tables \ref{tab:SAVEE_results} and \ref{tab:MELD_results}), the results of the multi-modal SER are significantly improved.
\begin{table}[t]
\centering
\captionsetup{font={small}}
\caption{The recognition results of the proposed FDAN in multi-dataset multi-modal SER tasks.}
\begin{tabular}{cp{2.2cm}<{\centering} cp{2.2cm}<{\centering} cp{2.2cm}<{\centering} cp{2.2cm}<{\centering}}
\hline
Tasks & WAR($\%$) & UAR($\%$) & w-F1($\%$)\\ \hline \hline
SAVEE$\{S_v, M_v, S_a\}$ & 87.50 & 87.14 & 87.70 \\
MELD$\{S_v, M_v, M_a\}$ & 81.78 & 78.31 & 81.01\\ \hline
\end{tabular}
\label{tab:cross-corpus}
\end{table}
\begin{figure}
  \centering
   \subfigure[SAVEE]{
    \includegraphics[width=4.8cm]{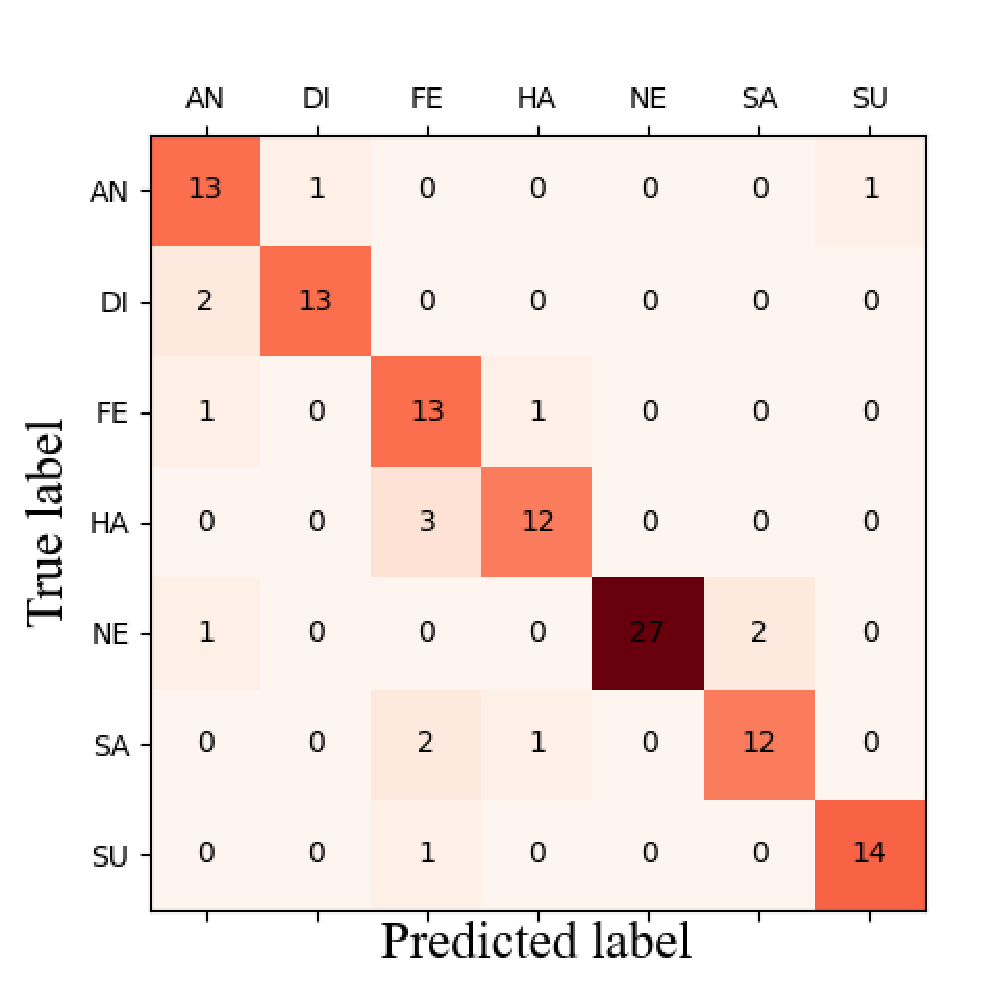}
    }
    \subfigure[MELD]{
    \includegraphics[width=4.8cm]{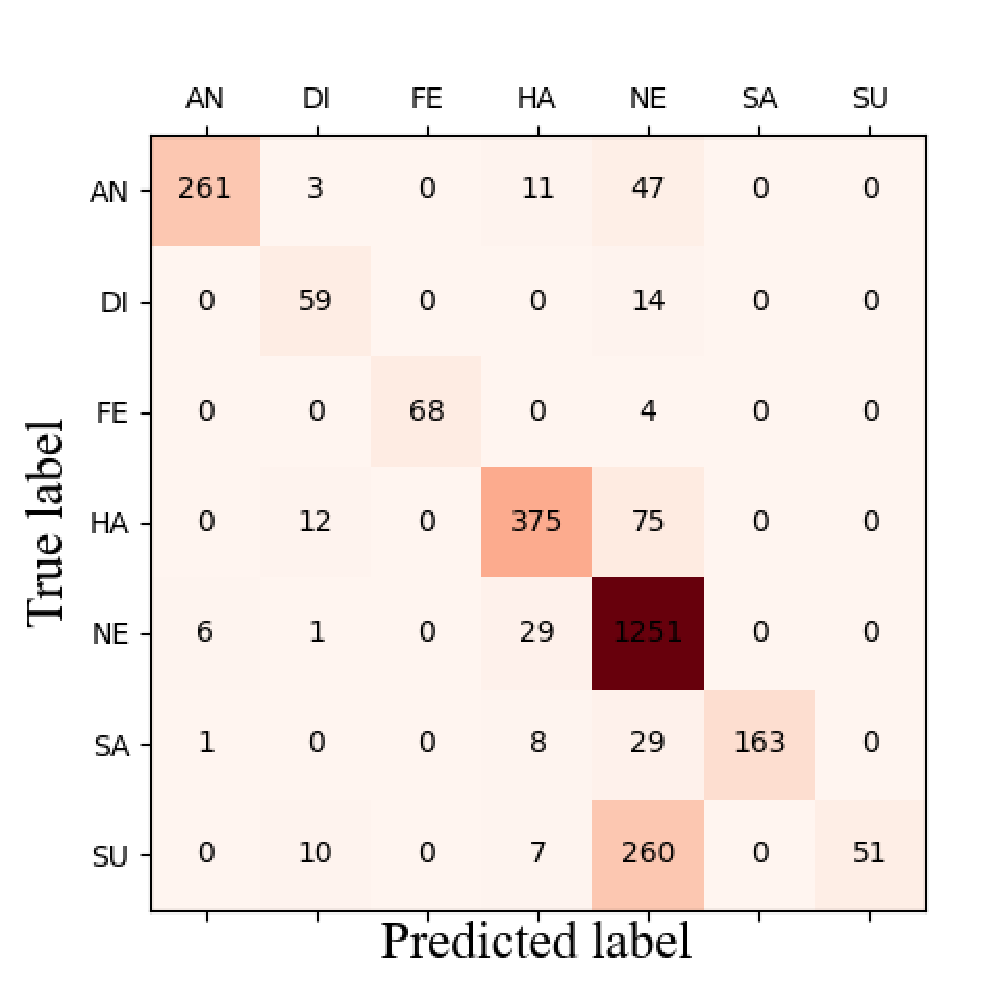}
    }
\captionsetup{font={small}}
\caption{Confusion matrices of our model. The horizontal axis represents the predicted label, and the vertical axis represents the true label (AN: anger, DI: disgust, FE: fear, HA: happiness, NE: neutral, SA: sadness, and SU: surprise).}
\label{fig:confusion}
\end{figure}
\subsection{Confusion matrices}
Fig. \ref{fig:confusion} shows the confusion matrices of the proposed method on the MELD and SAVEE datasets. As can be seen from the figure, first, the proposed method has a recognition accuracy of more than 75$\%$ for most emotion categories. Second, the SU, HA and SA emotions are easier to be confused with other emotions, and NE has the best recognition results. Finally, we can find that the recognition results of SU on MELD and SAVEE differ greatly, indicating that given an emotion category, the proposed model cannot well recognize the samples of each dataset.
\begin{figure}
  \centering
   \subfigure[SAVEE]{
    \includegraphics[width=5cm]{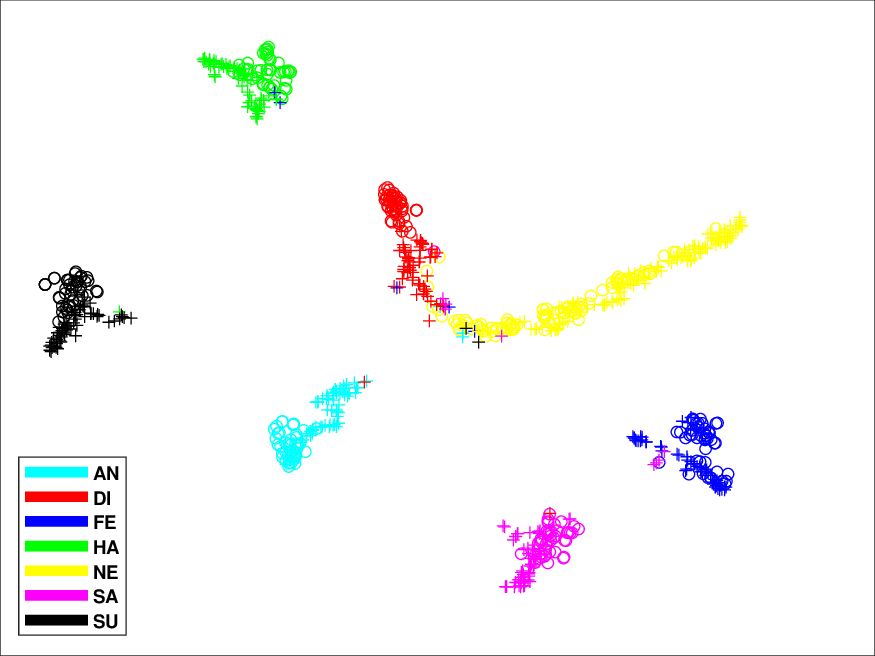}
    }
    \subfigure[MELD]{
    \includegraphics[width=5cm]{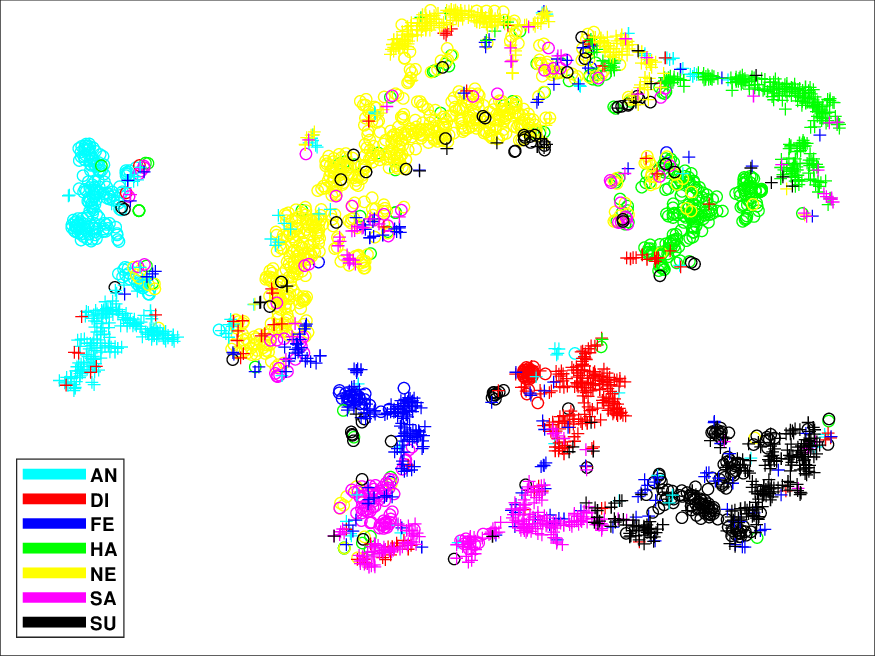}
    }
\captionsetup{font={small}}
\caption{The t-SNE data visualization results. The $\textbf{+}$ and \scalebox{1.2}{$\circ$} represent the visual and acoustic samples, respectively, and different colors represent different emotion categories (AN: anger, DI: disgust, FE: fear, HA: happiness, NE: neutral, SA: sadness, and SU: surprise).}
\label{fig:tsne}
\end{figure}
\subsection{t-SNE visualization}
\label{subsec:t-SNE visualization}
To better show the effectiveness of the proposed FDAN, we give the data visualization results using the t-SNE algorithm \cite{boureau2010learning}. Fig. \ref{fig:tsne} illustrates the t-SNE visualizations of visual and acoustic modal features obtained by FDAN on the SAVEE and MELD datasets. Since MELD is a large dataset, too many samples will destroy the visualization effects,  we randomly select 20$\%$ of the samples for visualization.\par
From the results in Fig. \ref{fig:tsne}, it can be seen that the proposed FDAN method effectively aligns the feature distributions of the visual and acoustic modalities and retains emotional discriminative information, resulting in a consistent emotional representation. These results sufficiently demonstrates the effectiveness of the proposed FDAN method.
\begin{table}[]
\centering
\captionsetup{font={small}}
\caption{The recognition results (WAR/UAR) ($\%$) of the proposed FDAN and its two special cases, i.e., FDAN$_1$ and FDAN$_2$.}
\begin{tabular}{cp{2.2cm}<{\centering} cp{2.2cm}<{\centering} cp{2.2cm}<{\centering} cp{2.2cm}<{\centering}}
\hline
Datasets & FDAN$_1$ & FDAN$_2$ & FDAN\\ \hline \hline
SAVEE & 83.33/ 82.91 & 79.63/ 79.25 & \textbf{86.66}/ \textbf{86.19} \\
MELD & 75.01/ 74.90 & 64.80/ 64.33 & \textbf{81.16}/ \textbf{75.91} \\ \hline
\end{tabular}
\label{tab:ablation}
\end{table}
\subsection{Ablation study}
To analyze the effectiveness, we further give the ablation study of FDAN. By removing the following components, i.e., cross-attention module and feature distribution adaptation, we can obtain two special cases of FDAN, i.e., FDAN$_1$ and FDAN$_2$. From the Table \ref{tab:ablation}, we have the following two observations. First, when the cross-attention mechanism is ignored, the recognition results of FDAN$_1$ decrease significantly. This result proves that the cross-attention module plays a positive role in our framework. Second, when setting the $\alpha =0$, the feature distribution adaptation is ignored, the recognition accuracy of FDAN$_2$ on all datasets decreases significantly. This proves that the feature distribution adaptation also plays a positive role in our framework. 
\section{Conclusions} 
\label{sec:conclusions}
In this paper, we propose a novel deep inductive transfer learning framework, called feature distribution adaptation network (FDAN), for multi-modal SER problem.  To the best of our knowledge, FDAN could be the first attempt to utilize deep transfer learning framework for multi-modal SER tasks. To demonstrate the effectiveness of our framework, extensive experiments are carried out on two popular datasets, and the results demonstrate that the proposed model achieves better performance than previous models. In the future, we would investigate to modify the proposed framework to tackle more challenging problems, such as the cross-corpus multi-modal SER.
%
%
%

\bibliographystyle{splncs04}
\bibliography{ref}
\end{document}